\documentclass[10pt, conference, compsoc]{IEEEtran}
\usepackage[noend]{algorithmic}
\usepackage{graphics}
\usepackage{graphicx}
\usepackage{mathrsfs}
\usepackage{amsmath}
\usepackage{amssymb}
\usepackage{amsfonts}

\begin{document}
%\title{Combining RRT and Stochastic Hill Climbing}
%\title{Combining Probabilistic Sampling Techniques and Stochastic Optimization for Path Planning}
%\title{A Multi-stage Probabilistic Algorithm for Dynamic Path-Planning}
\title{Combining a Probabilistic Sampling Technique and Simple Heuristics to
solve the Dynamic Path Planning Problem}


\author{\IEEEauthorblockN{Nicolas A. Barriga, Mauricio Solar}
\IEEEauthorblockA{Departamento de Inform\'atica\\
Universidad T\'ecnica Federico Santa Mar\'ia\\
Valparaiso, Chile\\
\texttt{\{nbarriga,msolar\}@inf.utfsm.cl}}
\and
\IEEEauthorblockN{Mauricio Araya-L\'opez}
\IEEEauthorblockA{MAIA Equipe\\
INRIA/LORIA\\
Nancy, France\\
\texttt{mauricio.araya@loria.fr}}
}


%Telephone: (800) 555--1212, Fax: (888) 555--1212}


\maketitle

\begin{abstract}
Probabilistic sampling methods have become very popular to solve single-shot
path planning problems. Rapidly-exploring Random Trees (RRTs) in particular have
been shown to be very efficient in solving high dimensional problems.
Even though several RRT variants have been proposed to tackle the dynamic
replanning problem, these methods only perform well in
environments with infrequent changes.
This paper addresses the dynamic path planning problem by combining simple
techniques in a multi-stage probabilistic algorithm. This algorithm uses
RRTs as an initial solution, informed local search to fix unfeasible paths
and a simple greedy optimizer. The algorithm is capable of recognizing when
the local search is stuck, and subsequently restart the RRT.
We show that this combination of simple techniques provides
better responses to a highly dynamic environment than the dynamic RRT variants.
\end{abstract}

\begin{IEEEkeywords}
artificial intelligence; motion planning; RRT; Multi-stage; local search; greedy
heuristics; probabilistic sampling;

\end{IEEEkeywords}

% \fi
\IEEEpeerreviewmaketitle

\section{Introduction}

The \emph{dynamic path-planning} problem consists in finding a suitable
plan for each new configuration of the environment by recomputing a
collision free path using the new information available at each time step~\cite{Hwang92}.
This kind of problem can be found for example by a robot trying to navigate
through an area crowded with people, such as a shopping mall or supermarket.
The problem has been addressed widely in its several flavors,
such as cellular decomposition of the configuration space~\cite{Stentz95},
partial environmental knowledge~\cite{Stentz94}, 
high-dimensional configuration spaces~\cite{Kavraki96}
or planning with non-holonomic constraints~\cite{Lavalle99}. 
However, simpler variations of this problem are complex enough that cannot be solved
with deterministic techniques, and therefore they are worthy to study.

This paper is focused on finding and traversing a collision-free path in
two dimensional space, for a
holonomic robot~\footnote{A holonomic robot is a robot in
which the controllable degrees of freedom is equal to the total degrees of
freedom.},
without kinodynamic restrictions~\footnote{Kinodynamic planning is a problem in which velocity and acceleration
bounds must be satisfied}, in two different scenarios:
\begin{itemize}
\item Dynamic environment: several unpredictably moving obstacles or adversaries. 
\item Partially known environment: some obstacles only become visible when
approached by the robot.
\end{itemize}
Besides from one (or few) new obstacle(s) in the second scenario we assume 
that we have perfect information of the environment at all times.

We will focus on continuous space algorithms and
won't consider
algorithms that use discretized representations of the configuration space, such
as D*~\cite{Stentz95},
because for high dimensional problems, the
configuration space becomes intractable in terms of both memory and computation
time, and there is the extra difficulty of calculating the discretization size,
trading off accuracy versus computational cost.

The offline RRT is efficient at finding solutions but they are far from
being optimal, and must be post-processed for shortening, smoothing or other qualities
that might be desirable in each particular problem. Furthermore, replanning RRTs
are costly in terms of computation time, as well as evolutionary and cell-decomposition 
approaches. Therefore, the novelty of this work is
the mixture of the feasibility benefits of the RRTs, the repairing capabilities
of local search, and the computational inexpensiveness of greedy algorithms,
into our lightweight multi-stage algorithm.

In the following sections, we present several path planning methods that can be
applied to the problem described above. In section \ref{sec:RRT} we review
the basic offline, single-query RRT, a probabilistic method that builds a
tree along the free configuration space until it reaches the goal state. 
Afterward, we introduce the most popular replanning variants of the
RRT: ERRT in section \ref{sec:ERRT}, DRRT in section \ref{sec:DRRT} and MP-RRT in section \ref{sec:MPRRT}.
Then, in section \ref{sec:hillclimbing} we present our new hybrid multi-stage algorithm with the
experimental results and comparisons in section \ref{sec:results}. Finally, the 
conclusions and further work are discussed in section \ref{sec:conclusions}.

\section{Previous and Related Work}
\label{sec:stateofart}
%TODO: Several algorithms has been proposed... blah blah !

\subsection{Rapidly-Exploring Random Tree}\label{sec:RRT}

One of the most successful probabilistic sampling methods for offline path planning
currently in use, is the Rapidly-exploring Random Tree (RRT), a single-query planner for
static environments, first introduced in
\cite{Lavalle98}. RRTs work towards finding a continuous path from a state
$q_{init}$ to a state $q_{goal}$ in the free configuration space $C_{free}$, by
building a tree rooted at $q_{init}$. A new state $q_{rand}$ is uniformly
sampled at random from the configuration space $C$. Then the nearest node,
$q_{near}$, in
the tree is located, and if $q_{rand}$ and the shortest path from $q_{rand}$ to
$q_{near}$ are in $C_{free}$, then $q_{rand}$ is added to the tree.
%(Algorithm
%\ref{alg:buildrrt}). 
The tree
growth is stopped when a node is found near $q_{goal}$. To speed up convergence,
the search is usually biased to $q_{goal}$ with a small probability.\\
In \cite{Kuffner00}, two new features are added to RRTs. First, the EXTEND
function %(Algorithm \ref{alg:extend}) }
is introduced, which, instead of trying
to add directly $q_{rand}$ to
the tree, makes a motion towards $q_{rand}$ and tests for collisions.
%\renewcommand{\figurename}{Algorithm}
%\begin{figure}[ht!]
%    \caption{BuildRRT$(q_{init},q_{goal})$}
%    \label{alg:buildrrt}
%    \begin{algorithmic}[1]
%        \REQUIRE $T$: empty tree
%        \STATE $T$.init($q_{init}$)
%        \WHILE {Distance$(T,q_{goal})> threshold$}
%            \STATE $q_{rand} \leftarrow$ RandomConfig()
%            \STATE Extend($T,q_{rand}$)
%        \ENDWHILE
%        \RETURN $T$
%    \end{algorithmic}
%\end{figure}

%\begin{figure}[ht!]
%    \caption{Extend$(T,q)$}
%    \label{alg:extend}
%    \begin{algorithmic}[1]
%        \STATE $q_{near} \leftarrow$ NearestNeighbor$(q,T)$
%        \IF{NewConfig$(q,q_{near},q_{new})$}
%            \STATE $T.add\_vertex(q_{new})$
%            \STATE $T.add\_edge(q_{near},q_{new})$
%            \IF{$q_{new} = q$}
%                \RETURN Reached
%            \ELSE
%                \RETURN Advanced
%            \ENDIF
%        \ENDIF
%        \RETURN Trapped
%    \end{algorithmic}
%\end{figure}
Then a
greedier approach is introduced,
%(the CONNECT function, shown in algorithms
%\ref{alg:rrtconnectplanner} and
%\ref{alg:connect}) 
which repeats EXTEND until
an obstacle is reached. This ensures that most of the time, we
will be adding states to the tree, instead of just rejecting new random states.
The second extension is the use of two trees, rooted at $q_{init}$ and
$q_{goal}$, which are grown towards each other. This significantly decreases the
time needed to find a path.
%\begin{figure}[ht!]
%    \caption{RRTConnectPlanner$(q_{init},q_{goal})$}
%    \label{alg:rrtconnectplanner}
%    \begin{algorithmic}[1]
%        \REQUIRE $T_a$: tree rooted at $q_{init}$
%        \REQUIRE $T_b$: tree rooted at $q_{goal}$
%        \STATE $T_a.init(q_{init})$
%        \STATE $T_b.init(q_{goal})$
%        \FOR{$k=1$ to $K$}
%            \STATE $q_{rand}\leftarrow $RandomConfig()
%            \IF{\textbf{not} (Extend$(T_a,q_{rand})=Trapped)$}
%                \IF{Connect$(T_b,q_{new})=Reached$}
%                    \RETURN Path$(T_a,T_b)$
%                \ENDIF
%            \ENDIF
%            \STATE Swap$(T_a,T_b)$
%        \ENDFOR
%        \RETURN Failure
%    \end{algorithmic}
%\end{figure}

%\begin{figure}[ht!]
%    \caption{Connect$(T,q)$}
%    \label{alg:connect}
%    \begin{algorithmic}[1]
%        \REPEAT
%            \STATE $S \leftarrow $Extend$(T,q)$
%        \UNTIL{\textbf{not} $(S = Advanced)$}
%    \end{algorithmic}
%\end{figure}

%\section{Dynamic variants of the RRT}\label{sec:dynamic}
\subsection{ERRT}\label{sec:ERRT}
The execution extended RRT presented in \cite{Bruce02} introduces two
RRTs extensions to build an on-line planner: the \emph{waypoint cache} and 
the \emph{adaptive cost penalty search}, which improves re-planning efficiency 
and the quality of generated paths. 
The waypoint cache is implemented by keeping a constant
size array of states, and whenever a plan is found, all the states in the plan
are placed in the cache with random replacement. Then, when the tree is no
longer valid, a new tree must be grown, and there are three
possibilities for choosing a new target state. With probability
P[\textit{goal}], the goal is chosen as the target; With probability
P[\textit{waypoint}], a random waypoint is chosen, and with remaining
probability a uniform state is chosen as before. Values used in \cite{Bruce02}
are P[\textit{goal}]$=0.1$ and P[\textit{waypoint}]$=0.6$.\\
In the other extension --- the adaptive cost penalty search --- the planner dynamically
modifies a parameter $\beta$ to help it finding shorter paths. A value of $1$ for $\beta$ will
always extend from the root node, while a value of $0$ is equivalent to the
original algorithm. 
Unfortunately, the solution presented in \cite{Bruce02} lacks of
implementation details and experimental results on this extension.

\subsection{Dynamic RRT} \label{sec:DRRT}
The Dynamic Rapidly-exploring Random Tree (DRRT) described in \cite{Ferguson06} is a
probabilistic analog to the widely used D* family of algorithms. It works by
growing a tree from $q_{goal}$ to $q_{init}$.
%, as shown in algorithm
%\ref{alg:drrt}. 
The principal advantage is that the root of the tree does 
not have to be changed during the lifetime of the planning
and execution. 
Also, in some problem classes
the robot has limited range sensors,
thus moving obstacles (or new ones) 
are typically near the robot and not near
the goal.
%This has the advantage that the
%root of the tree does not have to be moved during the lifetime of the planning
%and execution, and that in some problem classes, the robot has limited range
%sensors, thus moving or newly appearing obstacles will be near the robot, not
%near the goal. 
In general, this strategy attempts to trim smaller branches and
farther away from the root. When new information concerning the configuration space is
received, the algorithm removes the newly-invalid branches of the
tree,
%(algorithms \ref{alg:invalidatenodes} and \ref{alg:trimrrt})
and
grows the remaining tree, focusing, with a certain probability(empirically tuned
to $0.4$ in \cite{Ferguson06}) to a vicinity of the recently trimmed branches,
by using the a similar structure to the waypoint
cache of the ERRT. 
%(algorithm \ref{alg:choosetarget})% 
In experimental results
DRRT vastly outperforms ERRT. %QUESTION: How this can be possible if there is NO information about the results in Bruce02???

\subsection{MP-RRT}\label{sec:MPRRT}
The Multipartite RRT presented in \cite{Zucker07} is another RRT variant which
supports planning in unknown or dynamic environments. The MP-RRT maintains a
forest $F$ of disconnected sub-trees which lie in $C_{free}$, but which are not
connected to the root node $q_{root}$ of $T$, the main tree. At the start of a
given planning iteration, any nodes of $T$ and $F$ which are no longer valid are
deleted, and any disconnected sub-trees which are created as a result are placed
into $F$.
%(as seen in algorithms \ref{alg:mprrtsearch} and \ref{alg:pruneandprepend}).
With given probabilities,
the algorithm tries to connect $T$ to a
new random state, to the goal state, or to the root of a tree in $F$.
%(algorithm
%\ref{alg:selectsample}). 
In \cite{Zucker07}, a simple greedy smoothing heuristic is used, that tries to
shorten paths by skipping intermediate nodes. The MP-RRT
is compared to an iterated RRT, ERRT and DRRT, in 2D, 3D and 4D problems, with
and without smoothing. For most of the experiments, MP-RRT modestly outperforms
the other algorithms, but in the 4D case with smoothing, the performance gap in
favor of MP-RRT is much larger. The authors explained this fact due to MP-RRT
being able to construct much more robust plans in the face of dynamic obstacle
motion. Another algorithm that utilizes the concept of forests is the
Reconfigurable Random Forests (RRF) presented in \cite{Li02}, but without the
success of MP-RRT.

\section{A Multi-stage Probabilistic Algorithm}\label{sec:hillclimbing}

In highly dynamic environments, with many (or a few but fast) relatively small 
moving obstacles, regrowing trees are pruned too fast, cutting away important
parts of the trees before they can be replaced. This reduce dramatically
the performance of the algorithms, making them unsuitable for these class
of problems.
We believe that a better performance could be obtained 
by slightly modifying a RRT solution using simple obstacle-avoidance
operations on the new colliding points of the path by informed local search. Then,
the path could be greedily optimized if the path has reached the feasibility condition.

\subsection{Problem Formulation}

At each time-step, the proposed problem could be defined as 
an optimization problem with satisfiability constraints.
Therefore, given a path our objective is to minimize an evaluation function 
(i.e. distance, time, or path-points), with the $C_{free}$ constraint.
Formally, let the path $\rho=p_1p_2\ldots p_n$ a sequence of points, where
$p_i \in \mathbb{R}^n$ a $n$-dimensional point ($p_1 = q_{init}, p_n = q_{goal}$), 
$O_t\in \mathcal{O} $ the set of obstacles positions
at time $t$, and $eval:\mathbb{R}^n \times \mathcal{O} \mapsto \mathbb{R}$
an evaluation function of the path depending on the object positions.
Then, our ideal objective is to obtain the optimum $\rho*$ path that
minimize our $eval$ function within a feasibility restriction in the form

\begin{equation}
\displaystyle\rho*=arg\min_{\rho}[eval(\rho,O_t)]  \textrm{ with }  feas(\rho,O_t) = C_{free}
\label{eq:problem}
\end{equation}

where $feas(\cdot,\cdot)$ is a \emph{feasibility} function that equals to $C_{free}$
iff the path $\rho$ is collision free for the obstacles $O_t$.
For simplicity, we use very naive $eval(\cdot,\cdot)$ and $feas(\cdot,\cdot)$
functions, but this could be extended easily to more complex evaluation and
feasibility functions. 
The used $feas(\rho,O_t)$ function assumes that the robot is a punctual
object (dimensionless) in the space, and therefore, if all segments $\overrightarrow{p_i p_{i+1}}$
of the path do not collide with any object $o_j \in O_t$, we say that the path
is in $C_{free}$.
The $eval(\rho,O_t)$ function will be the length of the path, i.e. the sum of the distances between
consecutive points. This could be easily changed to any metric such as the time
it would take to traverse this path, accounting for smoothness,
clearness or several other optimization criteria.

\subsection{A Multi-stage Probabilistic Strategy}

If solving equation \ref{eq:problem} is not a simple task in static environments,
solving dynamic versions turns out to be even more difficult. In dynamic path planning 
we cannot wait until reaching the optimal solution because we must deliver
a ``good enough'' plan within some time quantum. Then, a heuristic approach 
must be developed to tackle the on-line nature of the problem. The heuristic
algorithms presented in sections \ref{sec:ERRT}, \ref{sec:DRRT} and \ref{sec:MPRRT},
extend a method developed for static environments, which produce a poor response
to highly dynamic environments and an unwanted complexity of the algorithms.

We propose a multi-stage combination of three simple heuristic probabilistic techniques
to solve each part of the problem: feasibility, initial solution and optimization.

\begin{figure}[ht]
\begin{center}
\includegraphics[width=0.5\textwidth]{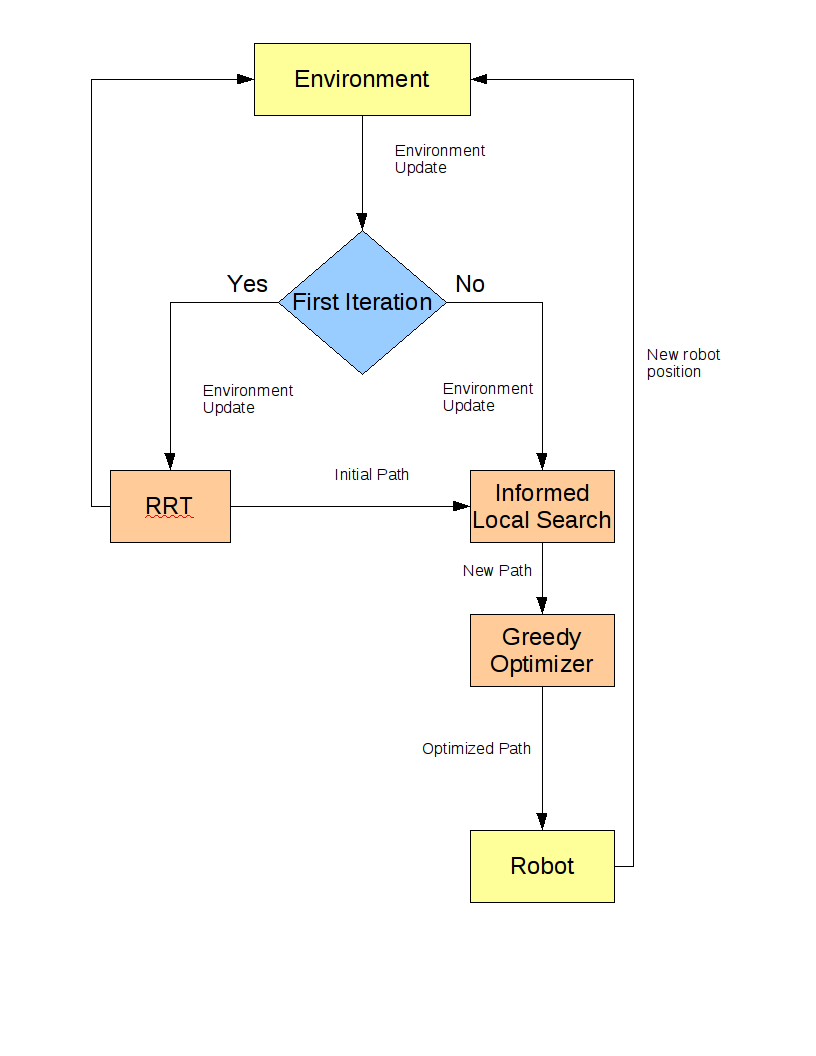}
\caption[\textbf{A Multi-stage Strategy for Dynamic Path Planning}]{\textbf{A Multi-stage Strategy for Dynamic Path Planning}. This figure describes
the life-cycle of the multi-stage algorithm presented here. The RRT, informed local search, and greedy heuristic are combined
to produce an expensiveness solution to the dynamic path planning problem.}
\label{fig:diag}
\end{center}
\end{figure}

\subsubsection{Feasibility}
The key point in this problem is the hard constraint in equation \ref{eq:problem}
which must be met before even thinking about optimizing. The problem is that in
highly dynamic environments a path turns rapidly from feasible to unfeasible
--- and the other way around --- even if our path does not change. 
We propose a simple \emph{informed local search} to obtain paths in $C_{free}$.
The idea is to randomly search for a $C_{free}$ path by modifying the nearest colliding
segment of the path. As we include in the search some knowledge of the problem,
the \emph{informed} term is coined to distinguish it from blind local search.
The details of the operators used for the modification of the path are described in
section \ref{sec:implementation}.
\subsubsection{Initial Solution}
The problem with local search algorithms is that they repair a solution that it is
assumed to be near the feasibility condition. 
Trying to produce feasible paths from scratch with local search (or even
with evolutionary algorithms~\cite{Xiao97}) is not a good idea due the randomness of the
initial solution. Therefore, we propose feeding the informed local search with a \emph{standard RRT} solution
at the start of the planning, as can be seen in figure~\ref{fig:diag}. 
\subsubsection{Optimization}
Without an optimization criteria, the path could grow infinitely large in time or
size. Therefore, the $eval(\cdot,\cdot)$ function must be minimized when a
(temporary) feasible path is obtained. A simple \emph{greedy} technique is used
here: we test each point in the solution to check if it can be removed
maintaining feasibility, if so, we remove it and check the following point,
continuing until reaching the last one.

\subsection{Algorithm Implementation}
\label{sec:implementation}

\newcounter{temp}
\newcounter{algo}

\setcounter{algo}{0}

\setcounter{temp}{\value{figure}}
\setcounter{figure}{\value{algo}}
\renewcommand{\figurename}{Algorithm}
\begin{figure}[ht]
    \caption{\bf Main()}
    \label{alg:main}
    \begin{algorithmic}[1]
        \REQUIRE $q_{robot} \leftarrow$ is the current robot position
        \REQUIRE $q_{goal} \leftarrow$ is the goal position
        \WHILE{$q_{robot} \neq q_{goal}$}
             \STATE {\bf updateWorld}$(time)$
             \STATE {\bf process}$(time)$
        \ENDWHILE
    \end{algorithmic}
\end{figure}
\renewcommand{\figurename}{Figure}
\addtocounter{algo}{1}
\setcounter{figure}{\value{temp}}

The multi-stage algorithm proposed in this paper works by alternating environment
updates and path planning, as can be seen in 
Algorithm~\ref{alg:main}. The first stage of the path planning (see Algorithm \ref{alg:process})
is to find an initial path using a RRT technique, ignoring any cuts that might happen during
environment updates. Thus, the RRT ensures that the path found 
does not collide with static obstacles, but might collide with dynamic obstacles in the future. 
When a first path is found,
the navigation is done by alternating a simple informed local search and 
a simple greedy heuristic as is shown in Figure~\ref{fig:diag}.

\setcounter{temp}{\value{figure}}
\setcounter{figure}{\value{algo}}
\renewcommand{\figurename}{Algorithm}
\begin{figure}[ht]
    \caption{\bf process$(time)$}
    \label{alg:process}
    \begin{algorithmic}[1]
        \REQUIRE $q_{robot} \leftarrow$ is the current robot position
        \REQUIRE $q_{start} \leftarrow$ is the starting position
        \REQUIRE $q_{goal} \leftarrow$ is the goal position
        \REQUIRE $T_{init} \leftarrow$ is the tree rooted at the robot position
        \REQUIRE $T_{goal} \leftarrow$ is the tree rooted at the goal position
        \REQUIRE $path \leftarrow$ is the path extracted from the merged RRTs
        \STATE $q_{robot} \leftarrow q_{start}$
        \STATE $T_{init}.init(q_{robot})$
        \STATE $T_{goal}.init(q_{goal})$
        \WHILE{time elapsed $<$ time}
            \IF{first path not found}
                \STATE {\bf RRT}$(T_{init},T_{goal})$
            \ELSE
                \IF{path is not collision free}
                    \STATE firstCol $\leftarrow$ collision point closest to robot
                    \STATE arc$(path, firstCol)$
                    \STATE mut$(path, firstCol)$
                \ENDIF
            \ENDIF
        \ENDWHILE
        \STATE {\bf postProcess}$(path)$
    \end{algorithmic}
\end{figure}
\addtocounter{algo}{1}
\renewcommand{\figurename}{Figure}
\setcounter{figure}{\value{temp}}

\begin{figure}[ht]
\begin{center}
\includegraphics[width=0.45\textwidth]{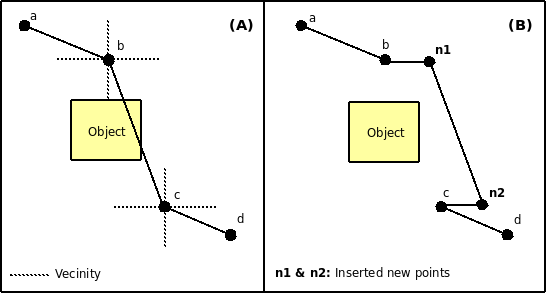}
\caption[\textbf{The arc operator}]{\textbf{The arc operator}. This operator draws an offset value $\Delta$ over a fixed interval called vicinity. 
Then, one of the two axises is selected to perform the arc and two new consecutive points are added to the path. 
$n_1$ is placed at a $\pm \Delta$ of the point $b$ and $n_2$ at $\pm \Delta$ of point $c$, both of them over the same selected axis. 
The axis, sign and value of $\Delta$ are chosen randomly from an uniform distribution.}
\label{fig:arc}
\end{center}
\end{figure}

\begin{figure}[ht]
\begin{center}
\includegraphics[width=0.45\textwidth]{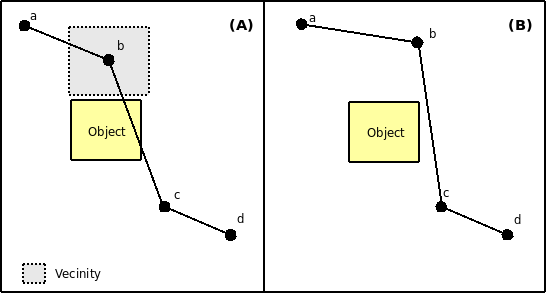}
\caption[\textbf{The mutation operator}]{\textbf{The mutation operator}. This operator draws two offset values $\Delta_x$ and $\Delta_y$ over a vicinity
region. Then the same point $b$ is moved in both axises from $b=[b_x,b_y]$ to $b'=[b_x \pm \Delta_x, b_y\pm \Delta_y]$, where the sign and offset values
are chosen randomly from an uniform distribution.}
\label{fig:mut}
\end{center}
\end{figure}

The second stage is the informed local search,
which is a two step function composed by the \emph{arc} and \emph{mutate} operators
(Algorithms \ref{alg:arc} and \ref{alg:mut}).
The first one tries to build a square arc around an
obstacle, by inserting two new points between two points in the path that form a
segment colliding with an obstacle, as is shown in Figure~\ref{fig:arc}. 
The second step in the function is a mutation
operator that moves a point close to an obstacle to a
random point in the vicinity, as is graphically explained in Figure~\ref{fig:mut}.
The mutation operator is inspired by the ones used in the Adaptive Evolutionary
Planner/Navigator(EP/N) presented in \cite{Xiao97}, while the arc operator is
derived from the arc operator in the Evolutionary Algorithm presented in
\cite{Alfaro05}.\\
Even though the local search usually produce good results for minor changes in the environment,
it does not when is faced to significant changes and is quite prone to getting stuck
in an obstacle. To overcome this limitation, our algorithm recognizes this situation, and restarts
an RRT from the current location, before continuing with the navigation phase.

\setcounter{temp}{\value{figure}}
\setcounter{figure}{\value{algo}}
\renewcommand{\figurename}{Algorithm}
\begin{figure}[ht]
    \caption{\bf arc$(path, firstCol)$}
    \label{alg:arc}
    \begin{algorithmic}[1]
        \REQUIRE vicinity $\leftarrow$ some vicinity size
        \STATE randDev $\leftarrow$ random$(-vicinity, vicinity)$
        \STATE point1 $\leftarrow$ path[firstCol]
        \STATE point2 $\leftarrow$ path[firstCol+1]
        \IF{random$() \% 2$}  
            \STATE newPoint1 $\leftarrow$ (point1[X]+randDev,point1[Y]) 
            \STATE newPoint2 $\leftarrow$ (point2[X]+randDev,point2[Y]) 
        \ELSE
            \STATE newPoint1 $\leftarrow$ (point1[X],point1[Y]+randDev) 
            \STATE newPoint2 $\leftarrow$ (point2[X],point2[Y]+randDev) 
        \ENDIF
        \IF{path segments point1-newPoint1-newPoint2-point2 are collision free}
            \STATE add new points between point1 and point2
        \ELSE
            \STATE drop new point2
        \ENDIF
    \end{algorithmic}
\end{figure}
\addtocounter{algo}{1}
\renewcommand{\figurename}{Figure}
\setcounter{figure}{\value{temp}}

\setcounter{temp}{\value{figure}}
\setcounter{figure}{\value{algo}}
\renewcommand{\figurename}{Algorithm}
\begin{figure}[ht]
    \caption{\bf mut$(path, firstCol)$}
    \label{alg:mut}
    \begin{algorithmic}[1]
        \REQUIRE vicinity $\leftarrow$ some vicinity size
        \STATE path[firstCol][X] $+=$ random$(-vicinity, vicinity)$
        \STATE path[firstCol][Y] $+=$ random$(-vicinity, vicinity)$
        \IF{path segments before and after path[firstCol] are collision free}
            \STATE accept new point
        \ELSE
            \STATE reject new point
        \ENDIF
    \end{algorithmic}
\end{figure}
\renewcommand{\figurename}{Figure}
\addtocounter{algo}{1}
\setcounter{figure}{\value{temp}}

The third and last stage is the greedy optimization heuristic,
which can be seen as a post-processing for path shortening, that
eliminates intermediate nodes if doing so does not create collisions,
as is described in the Algorithm \ref{alg:postProcess}.

\setcounter{temp}{\value{figure}}
\setcounter{figure}{\value{algo}}
\renewcommand{\figurename}{Algorithm}
\begin{figure}[ht]
    \caption{\bf postProcess$(path)$}
    \label{alg:postProcess}
    \begin{algorithmic}[1]
        \STATE i $\leftarrow$ 0
        \WHILE{i $<$ path.size$()$-2}
            \IF{segment path[i] to path[i+2] is collision free}
                \STATE delete path[i+1]
            \ELSE
                \STATE i $\leftarrow$ i+1
            \ENDIF
        \ENDWHILE
    \end{algorithmic}
\end{figure}
\renewcommand{\figurename}{Figure}
\addtocounter{algo}{1}
\setcounter{figure}{\value{temp}}

\section{Experiments and Results}\label{sec:results}

The multi-stage strategy proposed here has been developed
to navigate highly-dynamic environments, and therefore,
our experiments should be aimed towards that purpose.
Therefore, we have tested our algorithm in a highly-dynamic 
situation on two maps, shown in figures \ref{fig:office-dynamic} and
\ref{fig:800-dynamic}. For completeness sake, we have tested on the same two
maps, but modified to be a partially known
environment.
Also, we have ran the DRRT and MP-RRT algorithms over the same 
situations in order to compare the performance of our proposal.

\subsection{Experimental Setup}

\begin{figure}[ht]
\begin{center}
\includegraphics[width=0.45\textwidth]{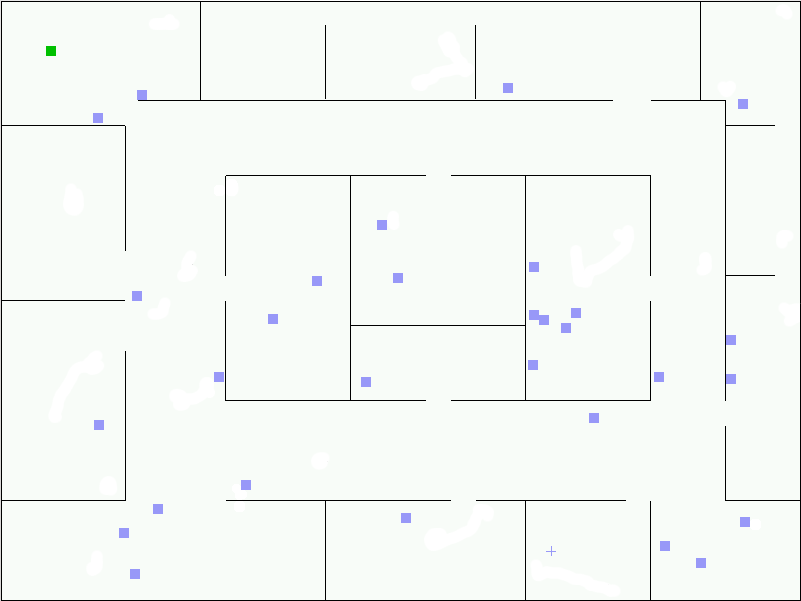}
\caption{The dynamic environment, Map 1. The \emph{green} square is our robot,
currently at the start position. The
\emph{blue} squares are the moving obstacles. The \emph{blue} cross is the goal.}
\label{fig:office-dynamic}
\end{center}
\end{figure}

\begin{figure}[ht]
\begin{center}
\includegraphics[width=0.45\textwidth]{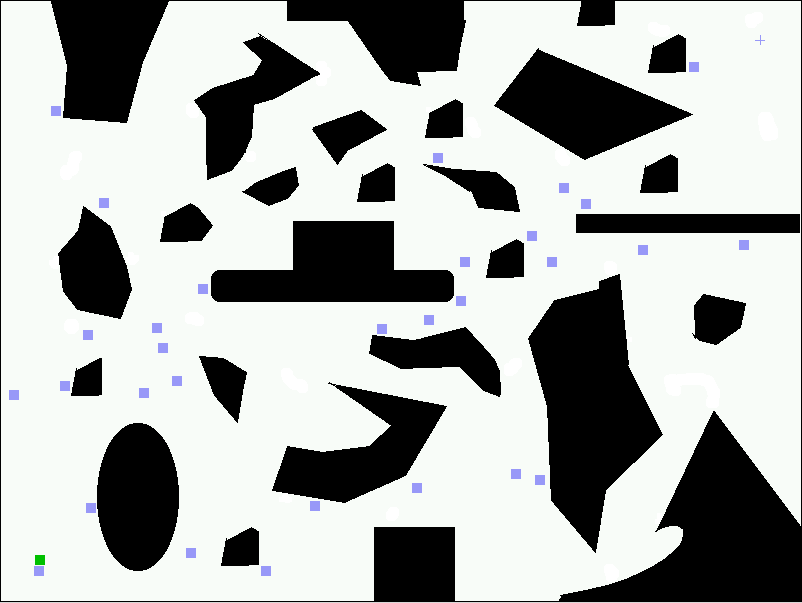}
\caption{The dynamic environment, Map 2. The \emph{green} square is our robot,
currently at the start position. The
\emph{blue} squares are the moving obstacles. The \emph{blue} cross is the goal.}
\label{fig:800-dynamic}
\end{center}
\end{figure}

The first environment for our experiments consists on two maps with 30 moving
obstacles the same size of the robot, with a random speed between 10\% and 55\%
the speed of the robot. This \emph{dynamic environments} are shown in figures
\ref{fig:office-dynamic} and \ref{fig:800-dynamic}.\\

\begin{figure}[ht]
\begin{center}
\includegraphics[width=0.45\textwidth]{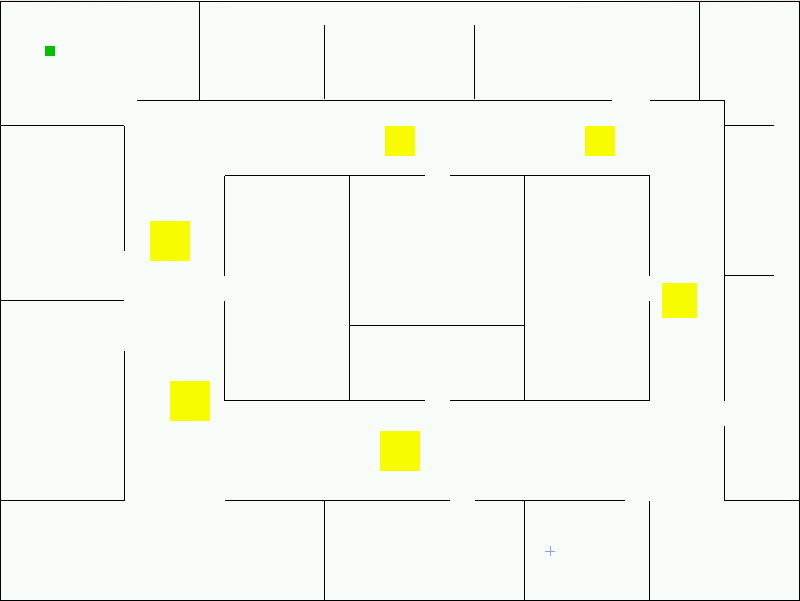}
\caption{The partially know environment, Map 1. The \emph{green} square is our robot,
currently at the start position. The \emph{yellow} squares are the suddenly 
appearing obstacles. The \emph{blue} cross is the goal.}
\label{fig:office-partial}
\end{center}
\end{figure}

\begin{figure}[ht]
\begin{center}
\includegraphics[width=0.45\textwidth]{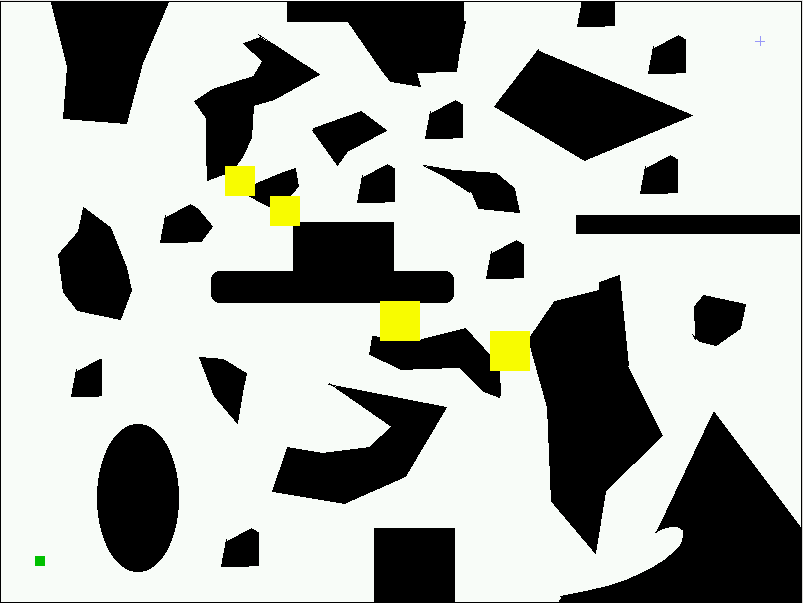}
\caption{The partially know environment, Map 2. The \emph{green} square is our robot,
currently at the start position. The \emph{yellow} squares are the suddenly 
appearing obstacles. The \emph{blue} cross is the goal.}
\label{fig:800-partial}
\end{center}
\end{figure}

The second environment uses the same maps, but with a few  obstacles, three to four
times the size of the robot, that become visible when the robot approaches each one of them. This
\emph{partially known environments} are shown in figure \ref{fig:office-partial}
and \ref{fig:800-partial}.\\

The three algorithms were ran a hundred times in each
environment.
The cutoff time was five minutes for all tests, after which, the robot was
considered not to have reached the goal. Results are presented concerning:
\begin{itemize}
\item {\it success rate:} the percentage of times the robot arrived to the goal
\item {\it number of nearest neighbor lookups performed by each algorithm(N.N.):} one of the
possible bottlenecks for tree-based algorithms
\item {\it number of collision checks performed(C.C.),} which in our specific
implementation takes a significant percentage of
the running time
\item {\it time} it took the robot to reach the goal
\end{itemize}

\subsection{Implementation Details}

The algorithms where implemented in C++ using a framework
\footnote{MoPa homepage: https://csrg.inf.utfsm.cl/twiki4/bin/view/CSRG/MoPa}
 developed by the same authors.\\

There are several variations that can be found in the literature when 
implementing RRTs. For all our RRT variants, the following are the details on
where we departed from the basics:
\begin{itemize}
\item We always use two trees rooted at $q_{init}$ and $q_{goal}$.
\item Our EXTEND function, if the point cannot be added without collisions to a
tree, adds the mid point between the nearest tree node and the nearest collision
point to it.
\item In each iteration, we try to add the new randomly generated point to both
trees, and if successful in both, the trees are merged, as proposed in
\cite{Kuffner00}.
\item We found that there are significant performance differences with allowing
or not the robot to
advance towards the node nearest to the goal when the trees are disconnected, as
proposed in \cite{Zucker07}. The problem is that the robot would become stuck
if it enters a small concave zone of the environment(like a room in a building)
while there are moving obstacles inside that zone, but otherwise it can lead to
    better performance. Therefore we present results for both kinds of behavior:
    DRRT-adv and MPRRT-adv, moves even when the trees are disconnected, while
    DRRT-noadv and MPRRT-noadv only
moves when the trees are connected.
\end{itemize}
In MP-RRT, the forest was handled simply replacing the oldest tree in it if the
forest had reached the maximum size allowed.

Concerning the parameter selection, the probability for selecting a point in the
vicinity of a point in the waypoint cache in DRRT was set to 0.4 as suggested in 
\cite{Ferguson06}. The probability for trying to reuse a sub tree in 
MP-RRT was set to 0.1 as suggested in \cite{Zucker07}. 
Also, the forest size was set to 25 and the minimum size of a
tree to be saved in the forest was set to 5 nodes.

\subsection{Dynamic Environment Results}
The results in tables \ref{table:dynamic1} and \ref{table:dynamic2} show that it takes 
our algorithm considerably less time than it takes the DRRT and MP-RRT to get to
the goal, with far less collision checks. It was expected that nearest neighbor
lookups would be much lower in the multi-stage algorithm than in the other two,
because they are only performed in the initial phase, not during navigation.

\begin{table}[ht]
\caption{Dynamic Environment Results, Map 1.}
\label{table:dynamic1}
\centering
\begin{tabular}{|c||c|c|c|c|}
\hline
Algorithm & Success \% & C.C. & N.N. & Time[s]\\
\hline
Multi-stage & 99 & 23502 & 1122 & 6.62 \\ 
\hline
DRRT-noadv & 100 & 91644 & 4609 & 20.57 \\
\hline
DRRT-adv & 98 & 107225 & 5961 & 23.72 \\
\hline
MP-RRT-noadv & 100 & 97228 & 4563 & 22.18\\
\hline
MP-RRT-adv & 94 & 118799 & 6223 & 26.86\\
\hline
\end{tabular}
\end{table}

\begin{table}[ht]
\caption{Dynamic Environment Results, Map 2.}
\label{table:dynamic2}
\centering
\begin{tabular}{|c||c|c|c|c|}
\hline
Algorithm & Success \% & C.C. & N.N. & Time[s]\\
\hline
Multi-stage & 100 & 10318 & 563 & 8.05\\ 
\hline
DRRT-noadv & 99 & 134091 & 4134 & 69.32\\
\hline
DRRT-adv & 100 & 34051 & 2090 & 18.94\\
\hline
MP-RRT-noadv & 100 & 122964 & 4811 & 67.26\\
\hline
MP-RRT-adv & 100 & 25837 & 2138 & 16.34\\
\hline
\end{tabular}
\end{table}

\subsection{Partially Known Environment Results}
The results in tables \ref{table:partial1} and \ref{table:partial2} show that our multi-stage
algorithm, although designed for dynamic environments, is also faster than the
other two in a partially known environment, though not as much as in the
previous cases.

\begin{table}[ht]
\caption{Partially Known Environment Results, Map 1.}
\label{table:partial1}
\centering
\begin{tabular}{|c||c|c|c|c|}
\hline
Algorithm & Success \% & C.C. & N.N. & Time[s]\\
\hline
Multi-stage & 100 & 12204 & 1225 & 7.96\\ 
\hline
DRRT-noadv & 100 & 37618 & 1212 & 11.66\\
\hline
DRRT-adv & 99 & 12131 & 967 & 8.26\\
\hline
MP-RRT-noadv & 99 & 49156 & 1336 & 13.82\\
\hline
MP-RRT-adv & 97 & 26565 & 1117 & 11.12\\
\hline
\end{tabular}
\end{table}

\begin{table}[ht]
\caption{Partially Known Environment Results, Map 2.}
\label{table:partial2}
\centering
\begin{tabular}{|c||c|c|c|c|}
\hline
Algorithm & Success \% & C.C. & N.N. & Time[s]\\
\hline
Multi-stage & 100 & 12388 & 1613 & 17.66\\ 
\hline
DRRT-noadv & 99 & 54159 & 1281 & 32.67\\
\hline
DRRT-adv & 100 & 53180 & 1612 & 32.54\\
\hline
MP-RRT-noadv & 100 & 48289 & 1607 & 30.64\\
\hline
MP-RRT-adv & 100 & 38901 & 1704 & 25.71\\
\hline
\end{tabular}
\end{table}

\section{Conclusions}\label{sec:conclusions}
The new multi-stage algorithm proposed here has a very good performance in
very dynamic environments. It behaves particularly well when several small
obstacles are moving around seemingly randomly. This is explained by the fact
that if the obstacles are constantly moving, they will sometimes move out of the
way by themselves, which our algorithm takes advantage of, but the RRT based
ones do not, they just drop branches of the tree, that could have been useful
again just a few moments later.\\
In partially known environments the multi-stage algorithm outperforms the RRT
variants, but the difference is not as much as in dynamic environments.
\subsection{Future Work}
There are several areas of improvement for the work presented in this paper.
The most promising seems to be to experiment with different on-line planners
such as the EP/N presented in \cite{Xiao97}, a version of the
EvP(\cite{Alfaro05} and \cite{Alfaro08}) modified to work in
continuous configuration space or a potential field navigator. Also, the local
search presented here, could benefit from the use of more sophisticated
operators.

Another area of research that could be tackled is extending this algorithm to
other types of environments, ranging from totally
known and very dynamic, to static partially known or unknown environments. An
extension to higher dimensional problems would be one logical way to go, as RRTs
are know to work well in higher dimensions.

Finally, as RRTs are suitable for kinodynamic planning, we only need to adapt
the on-line stage of the algorithm to have a new multi-stage planner for problem
with kinodynamic constraints.

\IEEEtriggeratref{3}


%\bibliographystyle{IEEEtran}
\bibliography{IEEEabrv,../../biblio.bib}

\begin{thebibliography}{10}

\bibitem{Alfaro05}
T.~Alfaro and M.~Riff.
\newblock {An On-the-fly Evolutionary Algorithm for Robot Motion Planning}.
\newblock {\em Lecture Notes in Computer Science}, 3637:119, 2005.

\bibitem{Alfaro08}
T.~Alfaro and M.~Riff.
\newblock {An Evolutionary Navigator for Autonomous Agents on Unknown
  Large-Scale Environments}.
\newblock {\em INTELLIGENT AUTOMATION AND SOFT COMPUTING}, 14(1):105, 2008.

\bibitem{Bruce02}
J.~Bruce and M.~Veloso.
\newblock Real-time randomized path planning for robot navigation.
\newblock {\em Intelligent Robots and System, 2002. IEEE/RSJ International
  Conference on}, 3:2383--2388 vol.3, 2002.

\bibitem{Ferguson06}
D.~Ferguson, N.~Kalra, and A.~Stentz.
\newblock Replanning with rrts.
\newblock {\em Robotics and Automation, 2006. ICRA 2006. Proceedings 2006 IEEE
  International Conference on}, pages 1243--1248, 15-19, 2006.

\bibitem{Hwang92}
Y.~K. Hwang and N.~Ahuja.
\newblock Gross motion planning---a survey.
\newblock {\em ACM Comput. Surv.}, 24(3):219--291, 1992.

\bibitem{Kavraki96}
L.~Kavraki, P.~Svestka, J.-C. Latombe, and M.~Overmars.
\newblock Probabilistic roadmaps for path planning in high-dimensional
  configuration spaces.
\newblock {\em Robotics and Automation, IEEE Transactions on}, 12(4):566--580,
  Aug 1996.

\bibitem{Kuffner00}
J.~Kuffner, J.J. and S.~LaValle.
\newblock Rrt-connect: An efficient approach to single-query path planning.
\newblock {\em Robotics and Automation, 2000. Proceedings. ICRA '00. IEEE
  International Conference on}, 2:995--1001 vol.2, 2000.

\bibitem{Lavalle99}
S.~LaValle and J.~Ku.
\newblock Randomized kinodynamic planning, 1999.

\bibitem{Lavalle98}
S.~M. Lavalle.
\newblock Rapidly-exploring random trees: A new tool for path planning.
\newblock Technical report, Computer Science Dept., Iowa State Univ., 1998.

\bibitem{Li02}
T.-Y. Li and Y.-C. Shie.
\newblock An incremental learning approach to motion planning with roadmap
  management.
\newblock {\em Robotics and Automation, 2002. Proceedings. ICRA '02. IEEE
  International Conference on}, 4:3411--3416 vol.4, 2002.

\bibitem{Stentz94}
A.~Stentz.
\newblock {Optimal and efficient path planning for
  partially-knownenvironments}.
\newblock In {\em 1994 IEEE International Conference on Robotics and
  Automation, 1994. Proceedings.}, pages 3310--3317, 1994.

\bibitem{Stentz95}
A.~Stentz.
\newblock {The Focussed D\^{}* Algorithm for Real-Time Replanning}.
\newblock In {\em International Joint Conference on Artificial Intelligence},
  volume~14, pages 1652--1659. LAWRENCE ERLBAUM ASSOCIATES LTD, 1995.

\bibitem{Xiao97}
J.~Xiao, Z.~Michalewicz, L.~Zhang, and K.~Trojanowski.
\newblock Adaptive evolutionary planner/navigator for mobile robots.
\newblock {\em Evolutionary Computation, IEEE Transactions on}, 1(1):18--28,
  Apr 1997.

\bibitem{Zucker07}
M.~Zucker, J.~Kuffner, and M.~Branicky.
\newblock Multipartite rrts for rapid replanning in dynamic environments.
\newblock {\em Robotics and Automation, 2007 IEEE International Conference on},
  pages 1603--1609, April 2007.

\end{thebibliography}
\bibliographystyle{abbrv}
%\begin{thebibliography}{1}


%\end{thebibliography}

\end{document}